\relax
\documentclass[letterpaper]{article} 
\usepackage{aaai22}  

\usepackage{times}  
\usepackage{helvet}  
\usepackage{courier}  
\usepackage[hyphens]{url}  
\usepackage{graphicx} 
\usepackage{natbib}  
\usepackage{caption} 
\DeclareCaptionStyle{ruled}{labelfont=normalfont,labelsep=colon,strut=off} 
\usepackage{booktabs}
\usepackage{algorithm}
\usepackage{algorithmic}
\usepackage{amsmath}

\usepackage{newfloat}
\usepackage{listings}
\floatstyle{ruled}
\newfloat{listing}{tb}{lst}{}
\floatname{listing}{Listing}

\title{Background Invariant Classification on Infrared Imagery by Data Efficient Training and Reducing Bias in CNNs}

\author{
    Maliha Arif,\textsuperscript{\rm 1,2}
    
    \thanks{Corresponding author}
    Calvin Yong,\textsuperscript{\rm 2}\thanks{This work was partially done when Calvin was part of CRCV lab.}
    Abhijit Mahalanobis\textsuperscript{\rm 1,2}\\
   
}

\affiliations{
    \textsuperscript{\rm 1}Center for Research in Computer Vision (CRCV)\\
    \textsuperscript{\rm 2}University of Central Florida\\

    4000 University Boulevard\\
    Orlando, Florida,USA 32816\\
    \{maliha.arif,calvinyong\}@knights.ucf.edu, amahalan@crcv.ucf.edu
}

\usepackage{bibentry}

\begin{document}

\maketitle

\begin{abstract}
Even though convolutional neural networks can classify objects in images very accurately, it is well known that the attention of the network may not always be on the semantically important regions of the scene. It has been observed that networks often learn background textures which are not relevant to the object of interest. In turn this makes the networks susceptible to variations and changes in the background which negatively affect their performance.  

We propose a new two-step training procedure called \textit{split training} to reduce this bias in CNNs on both Infrared imagery and RGB data. Our split training procedure has two steps: using MSE loss first train the layers of the network on images with background to match the activations of the same network when it is trained using images without background; then with these layers frozen, train the rest of the network with cross-entropy loss to classify the objects. Our training method outperforms the traditional training procedure in both a simple CNN architecture, and deep CNNs like VGG and Densenet which use lots of hardware resources, and learns to mimic human vision which focuses more on shape and structure than background with higher accuracy.

\end{abstract}

\begin{figure}[t]
\centering
\includegraphics[width=0.9\columnwidth]
{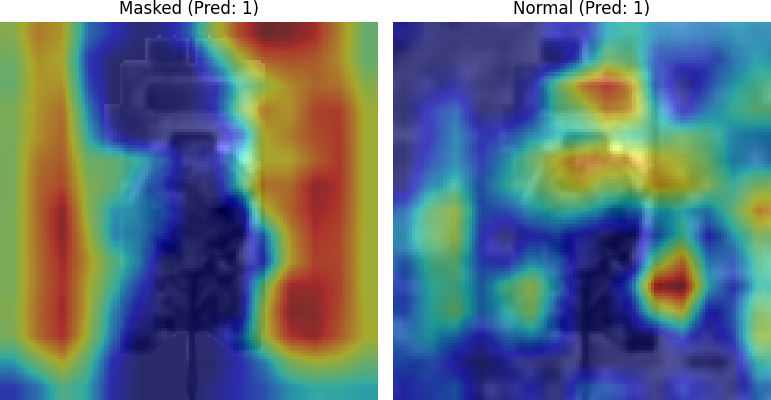} 
\caption{Grad-CAM/heatmaps obtained from a CNN model trained on targets(objects) with background.
Right: Unmasked - Original image of a tank(with background) and resultant Grad-CAM output during testing. Left:
Masked image (without background) and resultant Grad-CAM output at test time.}
\label{fig1}
\end{figure}

\section{Introduction}
Deep CNNs  with millions of parameters are able to learn complex patterns by developing low-level feature understanding leading to high-level scene understanding. However, training such deep networks requires huge amounts of training data that in turn requires too many computing resources. The very popular VGG \cite{simonyan2014very} was trained using 4 NVIDIA Titan Black GPUs,and it took 2–3 weeks to train the architecture on 1.3 million images of the ImageNet dataset. When the dataset is limited in size, the network is often unable to learn the semantically important regions of the scene for object classification. An example of this occurs for infrared object recognition where the data sets are relatively small, and applying transfer learning to networks trained with RGB data does not work well. Motivated by such issues,  we propose a split training method to efficiently learn from limited available data on infrared imagery and RGB data without relying on deep models, at the same time reducing background bias in CNNs. 

It is very important to understand what the learned representations in a deep CNN correspond to. In many cases, it is ignored as long as the accuracy and generalization capabilities of the network is observed to be high. We question this narrative in our research and try to see what the learned representations are using Grad-CAM \cite{selvaraju2017grad}. CNNs when trained with cross-entropy loss do not always use the actual target when making a prediction for an image; yet we would like such a model to be aware of the target. Figure \ref{fig1} shows the gradient class activation map (Grad-CAM) \cite{selvaraju2017grad} on test images from a deep CNN
model trained with images of a tank placed in a textured background.
On the right is a test image of a tank
with background. While the model has classified the
target correctly, the attention map shows the strong contribution of the background to the result. The test image on the left contains a tank but no background. Yet the same network (trained on images involving background), produces a Grad-CAM attention map which is largely outside the object of interest.  Our  goal is to avoid this problem by forcing the CNN to learn the semantic features (shape and structure) rather than background before it makes a prediction. 

\par
\begin{figure}
\centering
\includegraphics[width= 1.0\linewidth,height= 0.6\linewidth]{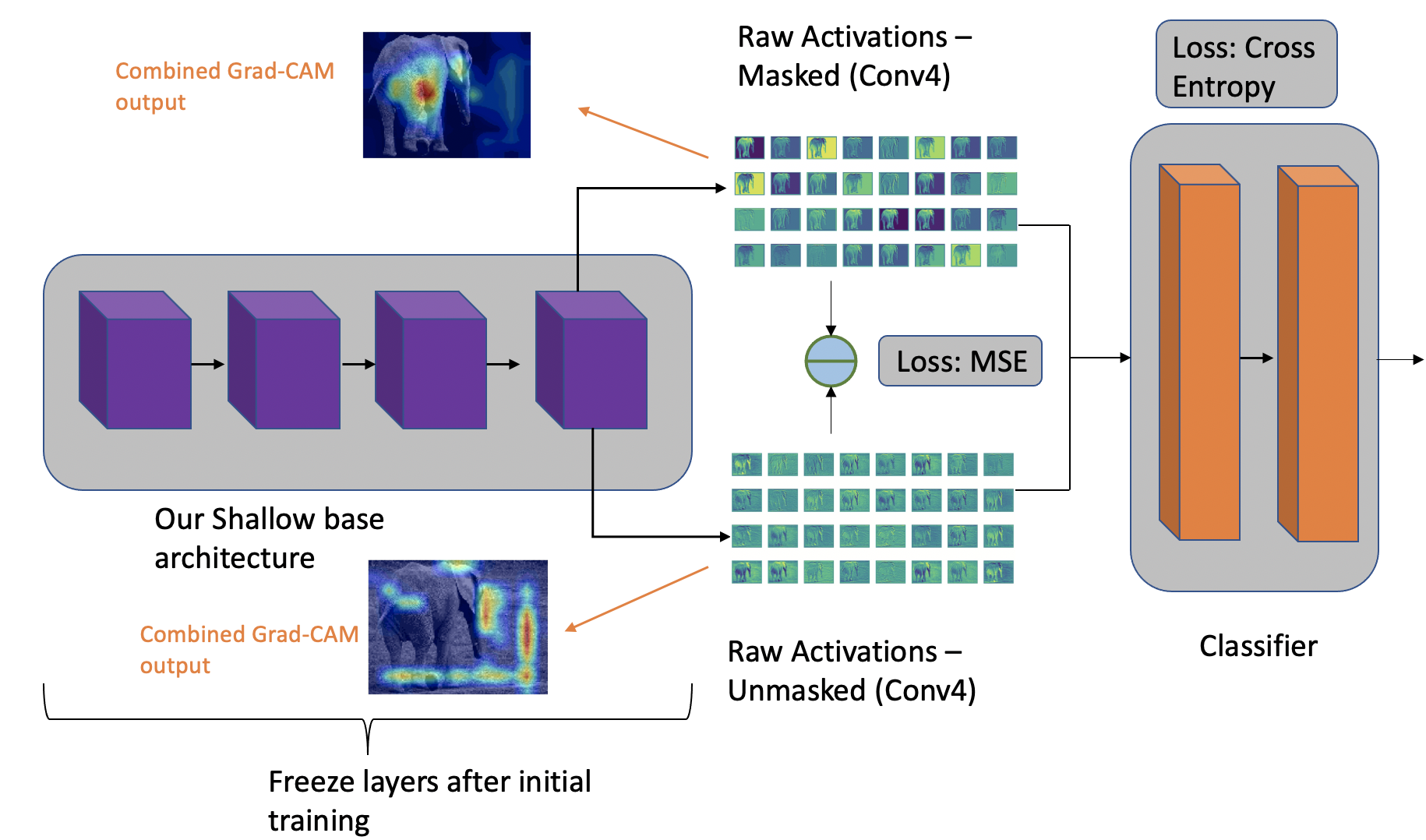}
\caption{Above, we show a 4 layered shallow network first trained on masked images to produce ground truth activations (top part). Another network using the same architecture is then trained using unmasked images(bottom part) and guided to match the ground truth activations using mean square error loss. The raw activations are combined to produce a Grad-CAM output for visualization purposes only.}
\label{BD}
\end{figure}

 Different works have tried to address the background bias problem. A very recent work by \cite{sehwag2020time} explicitly studies the effects of background on the classification performance of deep CNNs. Their focus is on background invariance and background influence , which they observe by switching backgrounds and by masking the foreground, and then performing classification respectively.  They show using a series of experiments on deep CNNs (like ResNet-18) and diverse datasets (like ImageNet), that even though the foreground is masked/removed, the network can still correctly identify the class. Similarly, when the background is switched and foreground remains intact, test accuracy decreases. On the VOC12 dataset \cite{everingham2015pascal} , test accuracy decreases from 75\% to 46\% when different backgrounds are introduced and object of interest or foreground is kept the same. So the question is, are the current training paradigms being used in deep learning the only way to train a deep CNN? What is the CNN actually using when it makes a prediction?  And how can available data be used efficiently for better test accuracy? To answer these questions, we propose a novel training strategy called ‘Split training’ which aims to teach the model to focus on the object (and ignore the background) by matching its activations at some specified layer to that of another model which was trained on masked (background removed) images and consequently produces \textit{ideal} activations as shown in Figure \ref{BD}. Our method outperforms the standard training method and transfer learning methods, and produces activation maps that show the object is used to make predictions rather than texture and background. In the following sections, we describe some related studies, our synthetic Infrared (grayscale) and RGB dataset which we use in our research, our proposed method and then various experimental results and ablation studies.

\section{Related Work}

Different works have tried to address the background bias problem. \cite{chen2017dual} propose a dual path network (DPN) which outperforms both ResNET and DenseNet deep CNNs. In particular, a shallow DPN with 26\% smaller model size surpasses ResNNXt-101 performance on ImageNet dataset which shows that just because a network is deep, it still might not be learning the best representations of the data. Further, it also uses 25\% less computational power and 8\% lower memory hence reducing the overall carbon footprint. \cite{xiao2020noise} study confirms that standard image classification models rely too much on signals from background and ignore foreground. Background correlations are largely predictive and influence model decisions to a great extent. Models often misclassify images even when correct foregrounds are present, up to 87.5\% of the time when random adversarial backgrounds are chosen. The backgrounds become a strong point of correlation between images and their labels which the authors show comprehensively via experiments involving ImageNet dataset with modified backgrounds. The backgrounds are created in a very interesting fashion. Some subsets include unmodified background but black foreground and some include unmodified background but tiled foreground. Some subsets also involve black background and unmodified foreground which we will use in our proposed training method. \cite{zhu2016object} exploit the visual hints in an image by learning them explicitly and adding them in a conventionally trained CNN model to increase test classification. They also show through a series of experiments how deep CNNs achieve high performance by relying on visual contents of the whole image, mostly the background. By contrast humans tend to learn and classify objects based on foreground which is shown by conducting human recognition experiments on either pure background or pure foreground (created using bounding boxes). The results indicate that human beings are outperformed by networks trained on background but are able to beat deep CNNs when trained on foreground implying that deep CNNs fail to classify images based on object shape and structure. 
A recent work that indicates how CNNs are biased by texture is \cite{geirhos2018imagenet}. The authors demonstrate texture bias using all major deep CNN models and experiment with stylized versions of ImageNet dataset, similar to the way done in \cite{xiao2020noise}. They create grayscale versions of objects which contain both shape and texture, silhouette images where object outline is filled with black color, images with only edges and lastly images with contrasting shapes and textures example a cat with texture of an elephant. All popular deep CNN for image classification fail to recognize shape and do better with texture, VGG-16 gives 17.2\% accuracy using shape vs. 82.8\% when using texture; GoogleNet gives 31.2\% using shape vs. 68.8\% when using only texture and ResNet-50 gives  22.1\%  accuracy using shapes as compared to 77.9\% when using texture. Human observers do much better as expected with 95.9\% accuracy when shapes are clearly defined. These results confirm our theory and are a motivation behind our proposed training method which aims to guide a network to focus more on object shape than background and texture.
Lastly, as mentioned before deep networks like VGG-16 fail to work well with Infrared images due to different domain types. VGG-16, ResNet and Densenet have been trained on ImageNet(an RGB dataset) hence transfer learning does not work well when such networks are finetuned on infrared imagery. 

\begin{figure}
\centering
\includegraphics[width= 0.9\linewidth,height= 0.5\linewidth]{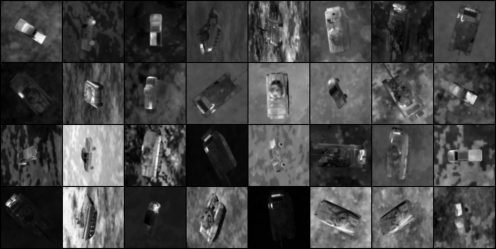}
\caption{Synthetic IR data samples of APC,tank and truck with background (Unmasked).}
\label{gendata1}
\end{figure}

\begin{figure}
\centering
\includegraphics[width= 0.9\linewidth,height= 0.5\linewidth]{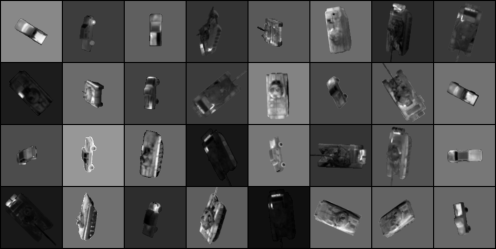}
\caption{Synthetic IR data samples of APC,tank and truck without background (Masked).}
\label{gendata2}
\end{figure}

\section{Datasets}

\textbf{Synthetic IR data - (Infrared Imagery): } The first dataset we use for our experiments is a synthetic infrared dataset of military vehicles. The dataset contains three
classes/targets: APC, tank, and truck. Each of the targets
are in 18 different backgrounds and thermal signatures,
and viewed in various angles, with azimuth angles between
0$^{\circ}$ and 359$^{\circ}$, and elevation angles of 15$^{\circ}$, 30$^{\circ}$, 45$^{\circ}$, 60$^{\circ}$,
75$^{\circ}$, and 90$^{\circ}$. This gives a total of 38880 images per class,
and thus a total of 116640 images for the entire dataset.
Examples from the dataset are shown in Figure \ref{gendata1}. Our dataset also contains masks that we can use to
remove the background of each target. Figure \ref{gendata2} shows
the same images as Figure \ref{gendata1}, but with the background
removed by using the mask. The masked dataset will
be used to train our primary model, while the dataset
including the background will be used to train the secondary
model.

\textbf{Preprocessing :} The infrared images have a different pixel range compared
to color(RGB) images. Infrared images use 16 bits per pixel,
and in the dataset, the pixel values range from
about 400 to 4000. For preprocessing, we centered the
image by subtracting the mean value
of the target from each pixel, then scaled the image to unit variance.
To generate the images with no background, we take
the preprocessed dataset with background, and set the
background pixels to 0. This ensures that the target
values are the same between the dataset with background
and the dataset without background.

\begin{figure}[ht]
\centering
\includegraphics[width= 1.0\linewidth,height= 0.6\linewidth]{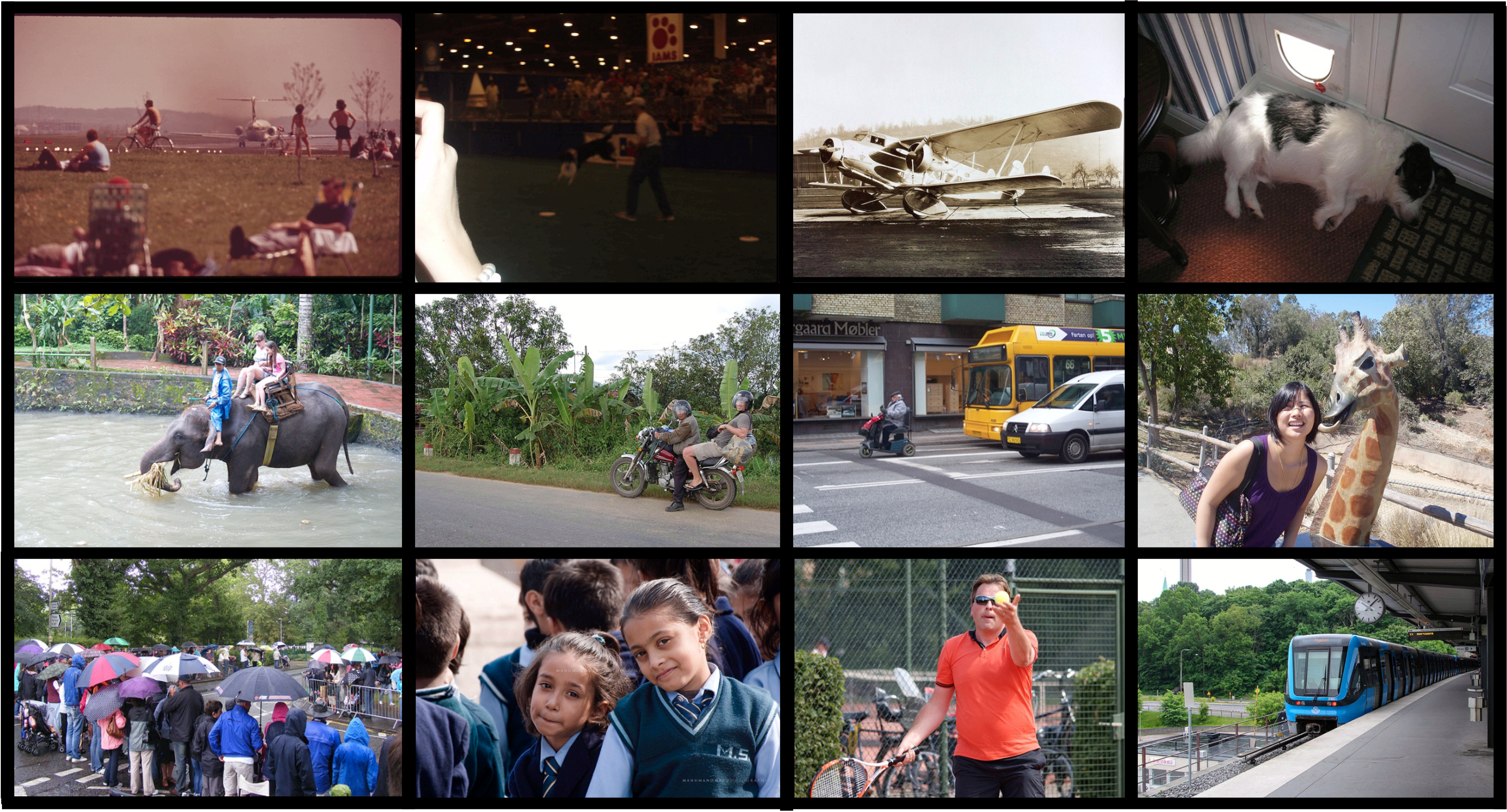}
\caption{MS-COCO sample images with background (Unmasked) - First row : Airplane(Left and second last to Right), Dog (Second to left and Right); 2nd row (L to R): Elephant, Motorcycle,Bus,Giraffe ; 3rd row (L to R): Umbrellas, Tie,Tennis racket,Clock.}
\label{coco1}
\end{figure}

\begin{figure}
\centering
\includegraphics[width= 1.0\linewidth,height= 0.6\linewidth]{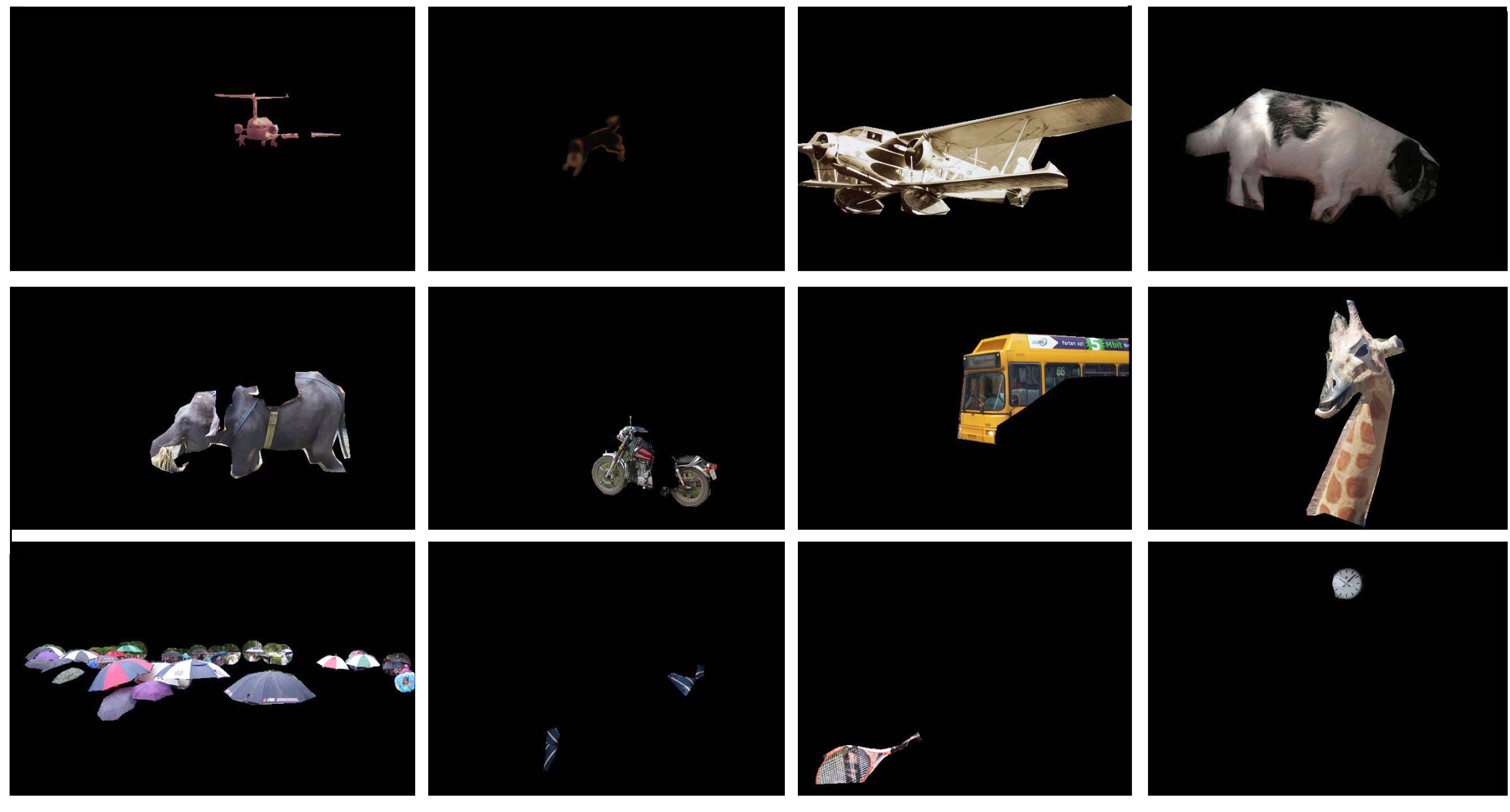}
\caption{MS-COCO sample images without background (Masked) - First row : Airplane(Left and second last to Right), Dog (Second to left and Right); 2nd row (L to R): Elephant, Motorcycle,Bus,Giraffe ; 3rd row (L to R): Umbrellas, Tie,Tennis racket,Clock.}
\label{coco2}
\end{figure}

\textbf{MS-COCO (RGB data) :} We also illustrate our work using RGB data to show our method's validity. MS-COCO is a main-stream computer vision dataset for object detection, segmentation and classification. Figure \ref{coco1} and \ref{coco2} show examples of unmasked and masked images which we use in our experiments. MS-COCO has 91 categories in all, with 81 suitable for segmentation. These object categories are grouped into 11 super categories. We make use of 10 of the 81 classes for instance segmentation and perform 2 experiments - one involving 3 classes and one involving 10. These classes include airplane, dogs, elephant, motorcycle, bus, giraffe, umbrella, tie, tennis racket and clock. In order to avoid confusion between classes, we selected training images which have only one object. However, we do end up with few images having multiple instances of the same class (and multiple classes) as can be seen in Figure \ref{coco1}.  We use a total of 20k training images (2k per class) and 1k test images. To create the set of masked images i.e. with background removed (Figure \ref{coco2} ), we use Pycoco \cite{pycoco} library and the ground truth annotations provided with the dataset. We mask the background by replacing the corresponding pixel values by zero. MS-COCO is a complex dataset with much of the images containing a lot of background, as a result of which the object of interest appears to be very small and not in the center. This becomes a good challenging point for our proposed method. 

\textbf{Preprocessing :} The RGB pixel values range from 0 to 255 unlike infrared images. We perform our experiments with and without mean target subtraction and find out that without mean target subtraction, our model performs better. Hence we stick to only normalizing our dataset by scaling the images to unit variance.  

\section{Method}

\begin{algorithm}[ht]
\caption{Split training method}
\label{alg:cap}
\begin{algorithmic}[1] 
\REQUIRE {$i \gets$ Training Images}
\FOR{$i$ $\epsilon$ \{ $ 1...n $ \} }
    \STATE $mean_ \gets i[mask!=0].mean()$ \COMMENT{For IR images}
    \STATE i -= mean                       \COMMENT{For IR images}
    \STATE normalize $i$
    \IF{masked}
        \STATE $i[mask==0] =0$
    \ENDIF
\ENDFOR
\STATE $m1 \gets$ Train primary model on masked images $i1$
\STATE $m2 \gets$ Train secondary model on Unmasked images $i2$
\STATE $k \gets$ Last feature layer of network
\FOR{$layer$ $\epsilon$ \{ $ 1...k $ \} }
    \STATE Optimize $layer$
    \STATE loss = MSE $[m1(i1) - m2(i2)]$ \COMMENT{For matching activations}
    \STATE use lr=1e-3
\ENDFOR
\FOR{$layer$ $\epsilon$ \{ $ 1...k+1...n $ \} }
    \STATE Load weights $(m2)$ and finetune layer $\epsilon$ \{$ 1...k $ \}
    \STATE Optimize $layer$
    \STATE loss = CE
    \STATE use lr=1e-4
\ENDFOR
\end{algorithmic}
\end{algorithm}

Our split training method comprises
of 3 main steps which are as follows:

\begin{enumerate}

    \item First train a primary model- m1 (a simple
CNN) using cross-entropy loss on
masked images (no background). 
    \item Next, train a secondary model- m2
(architecturally identical to the primary model) , using
unmasked images such that the MSE between the activations of one of the higher convolutional layers of the secondary and primary networks is minimized.
    \item Lastly, freeze the trained layers of the
secondary model. Using unmasked
images, train the remaining layers to
the end of the network using cross entropy (CE)
loss.
\end{enumerate}

The premise is that by design the primary network learns and responds to target information (since there is no background). Forcing the activations of the primary and secondary network to be similar encourages the latter to respond to the target and thereby learn to ignore the background.  The proposed approach is also shown in Algorithm \ref{alg:cap}. Our method differs from transfer learning methods in the following way. When finetuning a model pretrained on some dataset $D_1$ with another target dataset $D_2$, the convolutional layers are frozen with features learned from $D_1$ using large amounts of data, while the fully connected layers are finetuned to classify samples from $D_2$. In our split training method, all of the layers are trained using samples from the target dataset $D_2$. The model first starts with randomly initialized weights. Then the convolutional layers of the secondary model are trained to learn features from $D_2$ to match the activations of the primary model trained on $D_1$. Then, with those convolutional layers frozen, the fully connected layers are trained to learn weights to classify samples from $D_2$ given the features from the frozen convolutional layers.

\section{Experiments}

\begin{table*}[ht]
    \centering
    \caption{Test accuracies for each architecture and training method on synthetic IR (infrared) data. Numbers in parentheses are the standard deviation of the accuracies over 10 runs.}
    \label{tab:results}
\begin{tabular}{lllll}
    \toprule
    Architecture & Standard Training      & Finetuning / Transfer learning & Ours (last feature layer) & Ours (intermediate layer) \\
          \midrule
Simple    & 75.162\% (5.576\%)  & 69.160\% (3.050\%)     & \textbf{91.664\% (2.435\%)}   & 87.222\% (6.648\%) \\
Mobilenet & 73.580\% (8.243\%)  & 34.191\% (6.161\%)     & 74.319\% (5.536\%)   & \textbf{93.226\% (2.118\%)} \\
VGG11     & 72.798\% (13.000\%) & 48.224\% (7.565\%)     & \textbf{89.355\% (3.498\%) }  & 83.537\% (7.338\%) \\
Densenet  & 66.597\% (7.741\%)  & 50.098\% (4.361\%)     & 85.388\% (2.604\%)   & \textbf{88.557\% (3.712\%)} \\
\bottomrule
\end{tabular}
\end{table*}

\begin{table*}[ht]
    \centering
    \caption{Test accuracies on synthetic IR (infrared) data for each architecture and training method on the random background test dataset.}
    \label{tab:randbg}
\begin{tabular}{lllll}
  \toprule
Architecture    & Standard    & Finetuning / Transfer learning & Ours (last feature layer) & Ours (intermediate layer) \\
\midrule
Simple    & 70.417\% (5.830\%) & 83.875\% (2.372\%)    & \textbf{88.070\% (5.757\%) }     & 83.118\% (5.802\%) \\
Mobilenet & 67.558\% (6.048\%) & 45.961\% (8.007\%)    & 80.158\% (3.555\%)      & \textbf{95.969\% (2.189\%) }\\
VGG11     & 62.337\% (8.995\%) & 69.323\% (5.503\%)    & \textbf{86.238\% (4.417\%) }    & 80.548\% (5.469\%) \\
Densenet  & 66.378\% (1.363\%) & 55.573\% (1.698\%)    & 85.305\% (2.184\%)      & \textbf{85.753\% (2.208\%)} \\
\bottomrule
\end{tabular}
\end{table*}

\subsection{Synthetic IR data} 
We perform 2 sets of experiments using the synthetic IR data - one set of experiments use a train/test split involving ~120k original images using the data set as provided, and the second using images in random background. We use a simple CNN architecture which is a feedforward network with 4 Conv-BatchNorm-ReLU-MaxPool blocks. We train our network in 3 ways, one using masked images which gives us our ground truth activations, second using unmasked images and standard training procedure and then lastly, using unmasked images together with our split training method. For our split training method, we use the last feature layer as well as an intermediate layer for comparison.
We use Adam optimizer and learning rate of 1e-3 to train our initial layers using Mean Square Error(MSE) loss function. Once this part of the network is trained, we freeze the weights and train the top classifier part using Cross Entropy(CE) loss. We also compare our results with other architectures, details of which are ahead under section \textbf{Ablations and Comparisons}.
This is followed by another set of experiments where we modify our test set and make it more challenging.

Table \ref{tab:results} show the mean and standard deviations for
the accuracy of each model architecture trained with
various methods. Our method outperforms the standard (conventional)
training and finetuning methods, with performance
gains depending on the choice of feature layer
to match the secondary model’s activations with. With
Mobilenet, if we perform the MSE loss on the last
feature layer, we get that our method does as well as the
standard training method, yielding about 74\% accuracy.
We believe this is because the output shape at that layer
has 1280 channels each with size 4 x 4. The spatial
dimension at the output is very small, and the model may
not benefit from the information that the activations of
the secondary model give. When we use a layer closer to
the input, and where the shape is large enough, we get a
significant increase in performance, yielding a validation accuracy of 93\%. A poor choice in an intermediate
feature layer can be worse than the last feature layer, as
the results for VGG show.

\subsubsection{Random background Test Data}
\begin{figure}
\centering
\includegraphics[width= 1.0\linewidth,height= 1.0\linewidth]{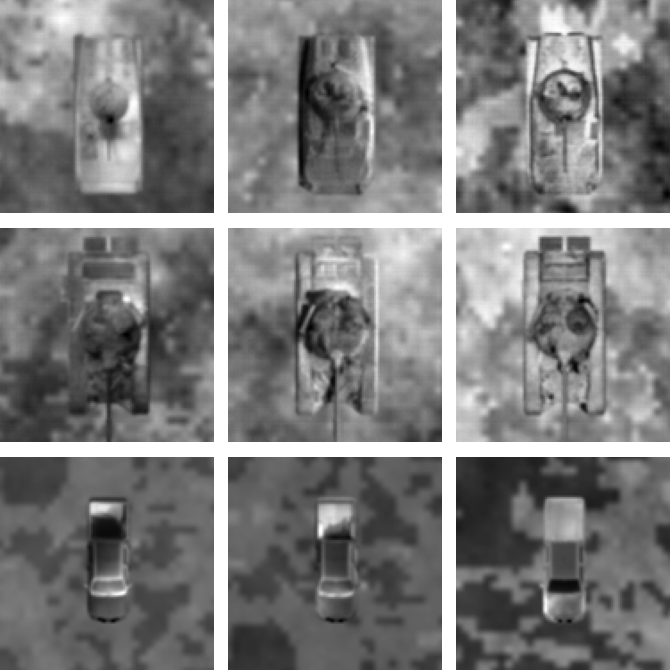}
\caption{Random background test set for Synthetic IR images.}
\label{val_set}
\end{figure}

For each class, we take 3 subsets of thermal signatures
to use for our test set. When choosing our subsets,
since some backgrounds may look similar, we visually
check them in order to avoid validating on images that may
look similar to the training set. This gives us about a 83-
17 training-test split. Figure \ref{val_set} shows the 9 subsets
of the data we choose, which are viewed with 0$^\circ$
azimuth and 90$^\circ$ elevation. From top to bottom, the rows correspond to  APC, tank, and
truck class.
Since our goal is to have the model focus on the target
and ignore the background, we also validate on the
synthetic IR dataset, by replacing the background with a random solid
color. That way, we hope that if our method does actually
learn to ignore the background, the model should also
do well on the dataset with random background.
We initialize each model
per training method with the same weights for uniformity. We find average of test accuracies on this random background data for each model across 10 runs, and report the mean and standard deviation in Table \ref{tab:randbg}.
Here we show the validation accuracy for the architectures
and training methods on the random background
test set. We see similar results to Table \ref{tab:results}. The finetuning
validation accuracy for the random background
test set is higher than the finetuning accuracy
on the regular validation set. This could be due to 
the pretrained model being trained on images with no
background and random test set (mix set) also containing images with no background(random solid color).
\subsection{Ablations and Comparisons}
\label{Ablations and Comparisons}
On synthetic IR dataset , we compare our method with the conventional training method using only cross entropy loss, and also transfer learning. For our method, we experimented with the last feature
layer for the MSE loss, along with an arbitrary chosen
layer between the input layer and the last feature layer. We benchmarked our method on 4 different models: a
simple CNN, Mobilenet \cite{sandler2018mobilenetv2}, VGG \cite{simonyan2014very}, and DenseNet \cite{huang2017densely}. To clarify, we only used the architecture of these deep models which are known to work really well on RGB data. When we say we did transfer learning, we trained the deep model first using masked images of our IR  dataset only and then finetuned using unmasked images and reported those results. We did not use pre-trained weights, trained on RGB - ImageNet data since they do not work well on Infrared imagery \cite{arif2021few}. Further, we do not want to employ a method that adds to the carbon footprint during training and uses lots of hardware resources/GPUs and memory.

\begin{table}[t]
    \centering
    \caption{Model performance on Mobilenet for varying output activation sizes.}
    \label{tab:layer}
    \begin{tabular}{lll}
    \toprule
    Activation Size & Validation Accuracy & Standard Deviation \\
    \midrule
    (24, 30, 30)   & 88.907\%     & 3.282\%   \\
    (32, 15, 15)   & 94.417\%     & 2.110\%   \\
    (64, 8, 8)     & 96.318\%     & 1.165\%   \\
    (96, 8, 8)     & 92.503\%     & 2.638\%   \\
    (1280, 4, 4)   & 93.226\%     & 2.118\%   \\
    \bottomrule
    \end{tabular}
\end{table}

\subsubsection{Choice of Intermediate Layer}
Table \ref{tab:layer} shows the validation accuracies for Mobilenet
on the infrared dataset with background for
varying output activation sizes. The activation
sizes are denoted by (c;h;w), where c is the number of channels,
h is the height, and w is the width. The table shows
that the intermediate feature layer that had an output
size of (64, 8, 8) yielded the highest accuracy. So in few cases, it could be 
beneficial to target an intermediate feature layer with a
large enough spacial size.

\textbf{Computation Time}: We also want to note the computation time for the different methods and training strategies. Table \ref{computation} shows these statistics. A simple network using our method takes less than 36000 s \textasciitilde  10 hrs to train on 3 classes of the infrared data. MobileNet is known to be nearly as accurate as VGG and is 32 times smaller and 27 times less computationally expensive \cite{sandler2018mobilenetv2}. VGG network alone took around 2-3 weeks when being trained on ImageNet, so whatever dataset it is finetuned on will cumulatively take longer than 1,209,600 s \textasciitilde  336 hours. In our exclusive case, since we did not use pretrained weights and trained on VGG11 architecture from scratch, the depth and additional parameters meant longer training time than 10 hrs.
Densenet networks are the most memory hungry networks among all popular deep models. Further, they are much deeper than VGG or MobileNet architectures. Computation time to train ImageNet dataset on Densenet is similar to VGG networks or more . Hence cumulative time to finetune on another dataset would be greater than 14 days \textasciitilde  1,209,600 s \textasciitilde  336 hours also. Training from scratch took longer than a simple CNN network that we proposed for our split training method. 

\begin{table}[t]
    \centering
    \caption{Total Computation (training) time using different pretrained networks and our method}
    \label{computation}
    \begin{tabular}{lll}
    \toprule
    Network     & Split Training/  & Training time(s)  \\
                & Transfer learning &              \\
    \midrule
    Simple      &  Split Training &   $<$ 36000 s        \\
    Mobilenet   &   Transfer learning &    $>$ 44800 s        \\
    VGG11       &   Transfer learning & $>$ 1,209,600 s  \\
    Densenet    &   Transfer learning  & $>$ 1,209,600 s   \\
    \bottomrule
    \end{tabular}
\end{table}

\subsection{MS-COCO}

On MS-COCO (RGB data) , we perform 2 experiments. First, we use only 3 classes as a starting point. We use 6k training images and 300 test images. We use a simple CNN as our shallow base network. It consists of 4 Conv-BatchNorm-ReLU-MaxPool blocks as also illustrated in Figure \ref{BD}. We first train this simple CNN using our masked images for 3 classes to obtain ground truth activations and also obtain the best model which focuses only on object since there is no background to get confused with. We use Adam optimizer with a learning rate of 1e-3 for the first 40 epochs and then 1e-4 for the last 20 epochs. We use a batch size of 32. We train it for 60 epochs. Since images in the dataset are of varying sizes , we resize all of them to 160 x 120 before training. To know what our baseline classification score is i.e. training via conventional method , we also train the same network using unmasked images.We keep all other hyperparameters the same for consistency. 
Lastly, we perform the experiment using our split training method. We first train the convolutional blocks using our proposed approach till the last feature layer and mean square error loss function. For best results, we train it for 30 epochs.We use Adam optimizer and learning rate of 1e-3.  Once this part is trained, we train the 2 classifier layers and finetune the bottom layers using a lower learning rate of 1e-4 for 30 more epochs. 
We perform a similar experiment involving 10 classes and use 20k training and 1k test images.  As shown in Table \ref{MS-COCO_results}, for both 3 and 10 class problems, our method's performance is better than \textit{ Standard(conventional)} training technique on unmasked images.

\begin{figure*}
\centering
\includegraphics[width= 1.0\linewidth,height= 0.3\linewidth]{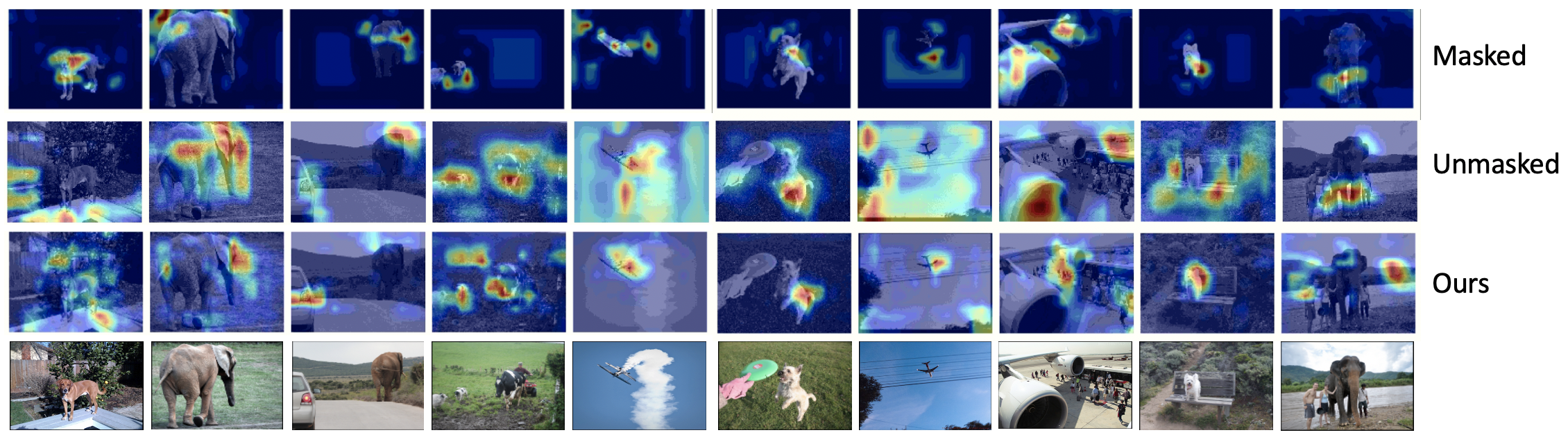}
\caption{Comparision of Grad-CAM heatmaps obtained using a CNN trained with our split
training method and conventional methods on masked and unmasked versions of MS-COCO dataset.}
\label{MS-coco_heatmaps}
\end{figure*}

\begin{table}[t]
    \centering
    \caption{Test accuracies on MS-COCO dataset}
    \label{MS-COCO_results}
    \begin{tabular}{llll}
    \toprule
    Classes & Masked  & Unmasked    & Ours (last \\
            &       &               & feature layer) \\
            & (Standard & (Standard & (Split  \\
            & Training) & Training) &  training) \\
    \midrule
    3       & 89.7\%   & 80.4\%     & \textbf{86.3\%}  \\
    10      & 73.13\%   & 58.01\%     & \textbf{65.61\%}  \\
    \bottomrule
    \end{tabular}
\end{table}

\subsection{Grad-CAM Results}

\begin{figure}
\centering
\includegraphics[width= 1.0\linewidth,height= 0.5\linewidth]{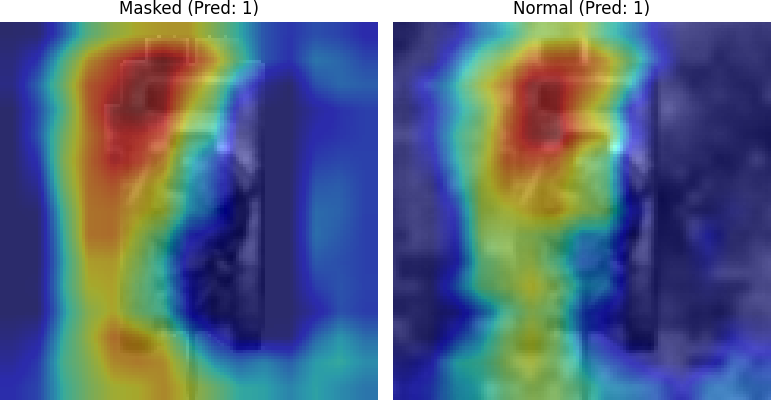}
\caption{Grad-CAM heatmaps from a CNN trained with our split
training method. L: Ground truth activations(heatmap) on masked image. R: Resultant heatmap after training on unmasked image.}
\label{Gendata_heatmaps}
\end{figure}

In the introduction section, we mentioned that the Grad-
CAM output from the model trained on images without
background yielded a more appealing output compared to
the Grad-CAM output from the model trained using background. Figure \ref{Gendata_heatmaps} shows the Grad-CAM outputs
for the model trained with our split training method on the infrared dataset. We
see that since our method trains the model to match the
activations of the primary model, the Grad-CAM/heatmaps closely match.
We also observe the method's performance on MS-COCO (RGB) data as demonstrated in Figure \ref{MS-coco_heatmaps}. Bottom row shows test images and top rows show the different heatmaps obtained at the last feature layer using different sets of input images (masked and unmasked).It is noteworthy that the attention maps produced by our method on test images with background (third row) are similar to the ideal attention maps obtained with masked images without background (top row). In contrast, the attention maps produced by conventional training techniques (second row) are observed to be less focused on the object of interest and semantically important parts of the scene.

\section{Conclusion}

We have shown that our split training method outperforms
the standard(conventional) training and finetuning
methods on Infrared and RGB data. When we split the training into two steps:
minimize the difference between the activations of
a primary model and a secondary model using MSE, and minimize the
final probabilities using CE loss, our model learns to focus on the
object, which is shown by the Grad-CAM outputs. For
our datasets, learning to focus on the object yielded better
classification performance subsequently.

\textbf{Acknowledgements} The authors would gratefully like to acknowledge the technical feedback of Dr.Nazanin Rahnavard \textit{(University of Central Florida)} while pursuing this research.


\bibliography{main.bib}

\end{document}